\documentclass[conference]{IEEEtran}

\usepackage{cite}
\usepackage{amsmath,amssymb,amsfonts}
\usepackage{graphicx}
\usepackage{textcomp}
\usepackage{xcolor}
\usepackage{booktabs}
\usepackage{multirow}
\usepackage{array}
\usepackage{url}
\usepackage{microtype}
\usepackage{xspace}

\newcommand{\method}{TRIM\xspace}          
\newcommand{\encoder}{TAP-Stats\xspace}    
\newcommand{\backbone}{TCN\xspace}

\newcommand{\etal}{\textit{et~al.}\xspace}

\begin{document}

\title{TRIM: Temporal Run-Time Monitoring of 3D Object Detection via Activation Statistics}

\author{
\IEEEauthorblockN{Hakan Yekta Yatbaz, Konstantinos Koufos, Mehrdad Dianati}
\IEEEauthorblockA{
School of Electronics, Electrical Engineering and Computer Science\\
Queen's University Belfast, Belfast, UK\\
}}

\IEEEoverridecommandlockouts
\IEEEpubid{\makebox[\columnwidth]{\parbox{\columnwidth}{\scriptsize 
\textcopyright~2026 IEEE. Personal use of this material is permitted. Permission from IEEE must be obtained for all other uses, in any current or future media, including reprinting/republishing this material for advertising or promotional purposes, creating new collective works, for resale or redistribution to servers or lists, or reuse of any copyrighted component of this work in other works.}\hfill} \hspace{\columnsep}\makebox[\columnwidth]{ }}

\maketitle

\begin{abstract}
   Reliable perception is essential to intelligent transportation systems, yet deep neural networks can silently miss safety-critical objects while continuing to produce apparently plausible outputs. Existing run-time monitors for 3D object detection generally assess each frame independently and therefore cannot exploit the temporal evolution of occlusion, point sparsity, and scene complexity. We present TRIM, a lightweight temporal monitor that detects missed-object failures without modifying or retraining the underlying detector. TRIM summarises intermediate activations using compact channel-wise statistics and processes short sequences of these summaries with a temporal convolutional network. Experiments on NuScenes show that temporal monitoring improves AUROC from \(0.686\) to \(0.704\) and reduces the false alarm rate at \(95\%\) recall from \(0.818\) to \(0.759\) compared with single-frame monitoring. Early and intermediate backbone features provide the strongest monitoring signal, while the temporal inference head requires only \(0.159\,\mathrm{ms}\) on a CPU.
\end{abstract}

\section{Introduction}
\label{sec:intro}
\IEEEpubidadjcol

Effective and reliable perception of the surrounding environment is
crucial to the safe operation of automated driving systems (ADS).
State-of-the-art systems employ LiDAR and detectors such as CenterPoint~\cite{yin2021center}
and PointPillars~\cite{lang2019pointpillars} to achieve high mean Average
Precision (mAP) on standard benchmarks. However, they remain vulnerable to
false negatives, including missed detections of vulnerable road users,
in challenging conditions such as sparse sensor returns, adverse weather,
or complex occlusion patterns~\cite{yatbaz2023survey}.
A pedestrian in the ego path emerging from behind a parked vehicle, or a cyclist near the sensor's field-of-view limit, may be missed entirely whilst the
detector otherwise appears to be performing reliably~\cite{yatbaz2025acm}. Such a failure mode is particularly concerning because it is silent; downstream planning receives no explicit indication that a relevant object has been missed.

A growing body of work on \textit{introspection}~\cite{yatbaz2023survey},
also referred to as self-assessment, integrity monitoring or run-time monitoring, seeks to
build auxiliary monitors that observe the detectors, specifically their internal states to flag degraded operation at run-time, without modifying or
retraining the detector itself.
Prior work has demonstrated that intermediate neural activation patterns
carry measurable signals of impending detection failures in both 2D
detectors~\cite{yatbaz2023iccv,yatbaz2024tiv} and, more recently, in
3D LiDAR-based detectors~\cite{yatbaz2024cvpr,yatbaz2024itsc,yatbaz2025acm}.
However, all existing monitors for LiDAR-based detection operate at the level of individual frames,
processing one point cloud (or a single camera frame) at a time and emitting a per-frame output, e.g., an alarm.
In practice, perception failures often develop gradually as occlusion increases, sensors degrade, or scenes become more complex. Tracking activations across frames may therefore reveal information that a single snapshot misses before a false negative occurs. We investigate whether sequences of compact
statistical summaries of intermediate detector activations contain
information associated with current or near-future false negatives, and
whether a temporal introspection model improves failure discrimination
over a matched frame-level baseline. Given the cost of full activation
maps in a temporal context, this preliminary study restricts itself to
simple activation statistics, evaluating \method{} as a low-cost
supervisory signal for edge or onboard perception pipelines rather than a
standalone safety-assurance mechanism.

\noindent 
To this end, the contributions of this work are:
\begin{enumerate}
    \item We formulate missed-object monitoring as a temporal representation-analysis problem and introduce TRIM, a run-time monitor for LiDAR-based 3D object detection that operates without modifying or retraining the underlying detector.

    \item Inspired by \cite{rahman2021perframe}, we develop TAP-Stats, a compact representation that summarises intermediate detector activations using five channel-wise statistics, enabling low-overhead temporal processing suitable for onboard or edge deployments.

    \item Evaluation on NuScenes shows temporal monitoring improves failure discrimination over an otherwise matched single-frame monitor. Temporal window length, prediction horizon, and activation layer selection are each ablated on the same test scenes across three seeds, and the false-alarm cost is reported across a sweep of target recalls rather than at a single operating point.
\end{enumerate}

\section{Related Work}
\label{sec:related}

\subsection{Run-time Monitoring of DNN-Based Perception}

Run-time self-assessment of DNN-based perception is also referred to as introspection or integrity monitoring~\cite{yatbaz2023survey,ferreira2024survey,schotschneider2025survey}. A comprehensive taxonomy in~\cite{yatbaz2023survey} organises methods into confidence-based, performance-based, inconsistency-based, and past-experience-based categories.

\textbf{Confidence-based methods} use low-confidence detector outputs as a
proxy for potential errors~\cite{feng2022review}, but DNN-based detectors
are known to produce high-confidence incorrect predictions, and a false
negative may leave no candidate detection at all, making confidence alone
insufficient for missed-object detection~\cite{yatbaz2023survey}.

\textbf{Performance-based methods} predict drops in aggregate metrics such
as mAP at deployment time. Rahman \etal~\cite{rahman2021online} predict per-frame mAP from internal detector features via a cascaded network, whilst their earlier
work~\cite{rahman2021perframe} showed that global pooling of
activation statistics is an effective proxy signal.
Yang \etal~\cite{yang2021introspective} instead predict, at the object
level, which specific detections a black-box detector is likely to miss.
Treating the problem as regression over an aggregate metric, however,
does not directly address the binary alert decision a run-time monitor must make.

\textbf{Inconsistency-based methods} exploit multi-modal or temporal
redundancy across sensors and detectors to flag
disagreements~\cite{antonante2023monitoring}, though deploying extra
sensors purely for monitoring is not always cost-effective in production
ADS.

\textbf{Activation-space methods} analyse the detector's intermediate representations directly. Class-conditional Gaussians fitted to layer-wise activations provide a strong out-of-distribution signal~\cite{lee2018mahalanobis}, a principle later adapted to object-detection monitoring. For 2D detectors, pruning, binarisation and scaling of activation patterns improve false-negative detection~\cite{yatbaz2023iccv,yatbaz2024tiv}. For 3D LiDAR detectors, early backbone layers prove more informative than final-layer features, their higher spatial resolution suiting the sparsity of point clouds~\cite{yatbaz2024cvpr}; spatial~\cite{yatbaz2024itsc} and object~\cite{yatbaz2025acm} filtering further focus monitoring on safety-critical regions and missed objects.

A shared limitation across all of the above is the frame-independence
assumption: each point cloud is monitored in isolation, without reference
to preceding frames. Among monitors for LiDAR-based 3D object detection, TRIM is, to our knowledge, the first to address this limitation directly.

\subsection{Temporal Context in Perception Monitoring}
Temporal context has been noted as an open direction in perception safety monitoring ~\cite{ferreira2024survey}. Rahman \etal~\cite{rahman2021online} use internal detector features in a cascaded network to estimate mAP over a sliding window of frames; TRIM instead evaluates causal sequences of compact channel-wise
activation statistics for binary missed-object monitoring in LiDAR-based 3D detection. Hecker \etal~\cite{hecker2018failure} predict imminent failures of an end-to-end driving-policy from camera frames, targeting the policy rather than a perception module, and operating on images rather than intermediate activations. To the best of our knowledge, this is the first study to evaluate causal temporal modelling of compact intermediate-activation statistics for false-negative monitoring in LiDAR-based 3D object detection.

\section{Method}
\label{sec:method}

\begin{figure*}[t]
  \centering
  \includegraphics[width=0.9\textwidth]{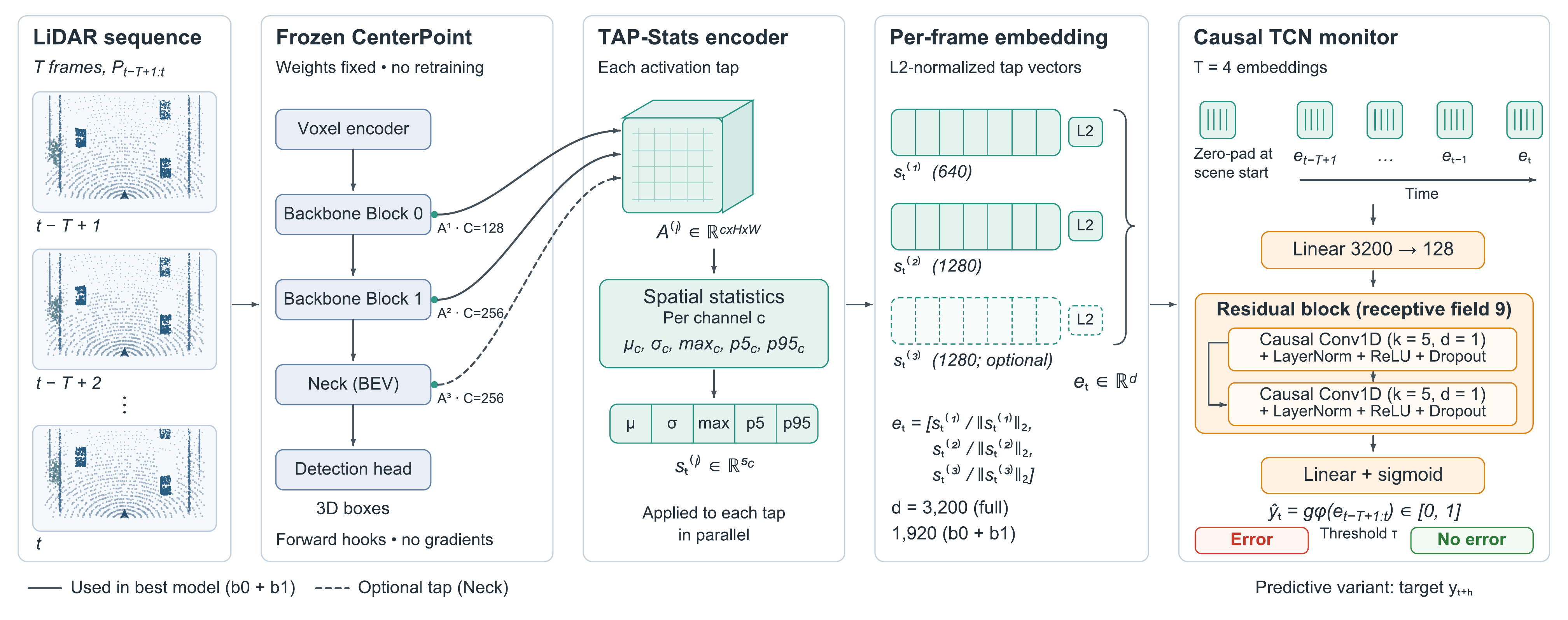}
  \caption{Overview of \method{}: Intermediate activations from a frozen CenterPoint detector are summarised by \encoder{} using five per-channel spatial statistics, then L2-normalised and concatenated into per-frame embeddings. A causal \backbone{} processes $T$ embeddings to predict the probability of missed detections at the target frame $t+h$, where $h=0$ denotes current-frame monitoring.}
  \label{fig:overview}
\end{figure*}

This section introduces \method{}, a temporal self-assessment
mechanism for LiDAR-based 3D object detection in ADS.
Unlike frame-level monitors, \method{} extracts compact statistical summaries from multiple
backbone layers of a fixed, pre-trained detector and reasons over
sequences of such summaries with a temporal model.
An overview is provided in Fig.~\ref{fig:overview}.

\subsection{Problem Formulation}
\label{sec:formulation}

Let $\mathcal{D} = \{(P_t, Y_t)\}_{t=1}^{T}$ denote a sequence of
point-cloud frames $P_t$ paired with ground-truth 3D bounding box sets $Y_t$.
A pre-trained detector $f_\theta$ produces predictions
$\hat{Y}_t = f_\theta(P_t)$.
A binary \textit{failure label} is defined as $y_t \in \{0,1\}$, where
$y_t=1$ (hereafter, ``Error'') if $f_\theta$ misses at least one
ground-truth object at frame $t$; that is, if there exists
$b \in Y_t$ with no class-consistent prediction in $\hat{Y}_t$ within the ground-plane
centre-distance threshold $\delta$. Otherwise, $y_t=0$ (hereafter,
``No-Error''). This convention follows prior introspection
work~\cite{yatbaz2025acm}.

The introspection task is to learn a monitor $g_\phi$ that, given access
to the internal activations of $f_\theta$ over a temporal window of
$T$ frames ending at time $t$, predicts $y_t$:
\begin{equation}
  \hat{y}_t =
  g_\phi\!\left(\mathbf{e}_{t-T+1},\ldots,\mathbf{e}_t\right),
  \label{eq:monitor}
\end{equation}
where $\mathbf{e}_t \in \mathbb{R}^d$ is a compact feature embedding
derived from the activations of $f_\theta$ at frame $t$.
The monitor is trained with focal loss using labels derived from
ground-truth annotations. At deployment, ground truth is unavailable and the
monitor operates from activations alone.

We additionally study a \textit{predictive} variant that targets
$y_{t+h}$ for a look-ahead horizon $h > 0$, enabling the monitor
to warn of impending failures. The case $h=0$ corresponds to the standard TRIM framework predicting false negatives in the current frame.

\subsection{TAP-Stats Encoder}
\label{sec:encoder}

The CenterPoint~\cite{yin2021center} detector is instrumented with three
Temporal Activation Pattern (TAP) hook points, selected to capture
information at different scales of the feature hierarchy:

\begin{itemize}
  \item \textbf{Block~0} (\texttt{pts\_backbone.blocks.0}):
    128 channels, capturing fine-grained spatial features early in the
    backbone before downsampling.
  \item \textbf{Block~1} (\texttt{pts\_backbone.blocks.1}):
    256 channels, capturing mid-level semantic features after one
    downsampling step.
  \item \textbf{Neck} (\texttt{pts\_neck.deblocks.1}):
    256 channels, capturing decoded spatial features after feature
    pyramid upsampling.
\end{itemize}

For each TAP point, forward hooks capture the output activation tensor
$A \in \mathbb{R}^{C \times H \times W}$ during inference without
introducing any gradient computation. Let $A_c \in \mathbb{R}^{H \times W}$ denote the $c$-th channel of $A$. 
The \encoder{} encoder computes five statistics over the spatial
dimensions $(H, W)$ for channel $c$:
\begin{equation}
  \mathbf{s}_c = \left[ \mu_c,\; \sigma_c,\; m_c,\; p_{5,c},\; p_{95,c} \right], \, c \in \{1,2, \ldots C\}, 
  \label{eq:tapstats}
\end{equation}
where $\mu_c$ and $\sigma_c$ are the spatial mean and standard deviation,
$m_c=\max(A_c)$ is the spatial maximum, and $p_{5,c}$, $p_{95,c}$ are the
5th and 95th percentiles of $A_c$.
This yields $5C$ values per TAP point, resulting in a per-frame embedding
of dimension
\begin{equation}
  d = 5 \times (128 + 256 + 256) = 3{,}200.
  \label{eq:dim}
\end{equation}
All embeddings are L2-normalised per layer prior to concatenation to
prevent any single TAP point from dominating the representation.
TAP-Stats encoding requires less than 1 ms per frame on an NVIDIA RTX 5090 (median over the test set) and requires no gradient computation.

\subsection{Temporal Convolutional Network Monitor}
\label{sec:tcn}

Given a sequence of embeddings $(\mathbf{e}_{t-T+1}, \ldots,
\mathbf{e}_t)$ of length $T$, the monitor $g_\phi$ is implemented as a
single-block TCN~\cite{bai2018tcn}.
Causal convolutions ensure that the monitor observes only past and
present frames, making it directly deployable in a streaming inference
setting.
The network architecture comprises: a linear projection layer reducing
the 3{,}200-dimensional input to a hidden dimension of 128; one causal
residual block of two convolutions with kernel size $k{=}5$ and dilation
$d{=}1$, each followed by layer normalisation, ReLU activation, and
dropout; and a binary classification head projecting to a scalar failure
probability via sigmoid.
Sequences are causal, scene-local, and left-padded with zero embeddings
at scene boundaries, so no information crosses between scenes.
With two convolutions, the receptive field of the final time step spans
nine frames, which covers every window length evaluated in this work.

Training uses focal loss~\cite{lin2017focal} to handle class imbalance
(13.5\% positive frames on the training split), with an Adam optimiser
and a class-balanced sampler.
Hyperparameters, including hidden dimension, kernel size, window size
$T$, and learning rate, were selected via Optuna Bayesian optimisation
on the internal validation split, and are held fixed across every configuration and
seed reported below except where explicitly ablated.

\subsection{Label Construction}
\label{sec:labels}

Following the centre-distance matching principle of the NuScenes evaluation protocol~\cite{caesar2020nuscenes}, ground-truth and predicted boxes are matched greedily by ground-plane centre distance. Ground-truth objects are restricted to the eight annotated road-user categories (car, truck, construction vehicle, bus, trailer, motorcycle, bicycle, pedestrian) and to the region $ 0 \leq x \leq 50$\,m, $|y| \leq 5.5$\,m, concentrating the label on the forward corridor relevant to the ego vehicle; predictions are drawn from the same categories above a detection score of $0.1$. A frame is labelled Error if at least one ground-truth object has no matching prediction within $\delta=3$\,m, else No-Error -- identifying whether a miss occurs, but neither localising it nor distinguishing a complete omission from a poorly localised prediction.

\section{Performance Evaluation }
\label{sec:experiments}

This section presents the experimental setup and performance evaluation
of \method{}.
The primary objective is to evaluate whether temporal reasoning over
activation statistics improves the detection of false negatives compared
to frame-level and single-layer baselines, and to quantify the benefit
of each design choice through systematic ablation.

\subsection{Experimental Setup}
\label{sec:setup}

\textbf{Dataset:}
We use NuScenes~\cite{caesar2020nuscenes}, which provides 1{,}000
driving scenes of 20\,s each, annotated at 2\,Hz across six camera
views and one 32-beam LiDAR; only LiDAR input is used. Scenes are
split at the scene level into a monitor-training partition (560
scenes, 22{,}507 frames, 3{,}040 Error) used for model training, an
internal validation partition (140 scenes, 5{,}623 frames, 924 Error)
used for hyperparameter optimisation and checkpoint selection, and a test set (the NuScenes validation split, nus\_val; 150 scenes, 6,019 frames, 918 Error, 15.25\%) used only for final reporting and never for checkpoint or threshold selection. No scene or sample appears in more than one partition. Every learned
result is reported over three fixed seeds (1024, 1025, 1026) as mean
$\pm$ population standard deviation.

\textbf{Detector:}
CenterPoint~\cite{yin2021center} is used as the base detector.
Detector weights are fixed throughout and no retraining or fine-tuning
is performed, consistent with the deployment-time scenario of \method{}.

\textbf{Evaluation metrics:}
Because the task is imbalanced, the main comparisons emphasise AUROC as the primary ranking metric.
Thresholded behaviour is reported at a fixed probability of $0.5$ through the Error recall $R_{+}$, the No-Error recall $R_{-}$, and the false positive rate $\mathrm{FPR}=1-R_{-}$. Reporting both class recalls, rather than accuracy or recall alone, is essential here: at a positive prevalence of 15.25\% a monitor can appear
to perform well on one class while failing badly on the other. High-recall operating behaviour is characterised by the False Alarm Rate at 95\% True Positive Rate (FAR@TPR95) and, in Section~\ref{sec:far_sweep}, by a sweep of FAR across nine target recalls.

\textbf{Baselines:}
Performance is compared against \textbf{Random} (AUROC $=0.500$), \textbf{Confidence score} (inverted maximum CenterPoint score), its four-frame \textbf{temporal} average, and \textbf{$k$-NN} ($k{=}50$), which uses the mean squared distance to the 50 nearest No-Error training embeddings in the 3,200-dimensional feature space. We also compare against \textbf{Frame-only} (TAP-Stats, $T{=}1$), which removes temporal context, and \textbf{Single-layer TAP} (Block~1 only), which removes multi-layer fusion.

\subsection{Test Set Performance}
\label{sec:main_results}

Table~\ref{tab:main_results} reports the main results. First, non-parametric k-NN retrieval over TAP-Stats embeddings (AUROC~$= 0.617$) already exceeds random indicating that the embeddings are informative even without training. Second, a trained frame-level model achieves AUROC~$=0.686$, and adding temporal context raises this to $0.704$; the two models share features, labels, scene split, seeds, and optimiser settings and differ only in the temporal window, so the gain can be attributed to temporal context. The same pattern holds for the full three-tap representation ($0.679 \to 0.692$). Third, the strongest representation is the lighter b0+b1 variant rather than the full b0+b1+neck fusion, leading on AUROC and FAR@TPR95 simultaneously.
\begin{table}[t]
  \centering
  \caption{Test set results (mean $\pm$ std over 3 seeds). Learned monitors use a 0.5 threshold; baseline thresholds maximise validation balanced accuracy. Best learned results in bold.}
  \label{tab:main_results}
  \scriptsize
  \setlength{\tabcolsep}{2pt}
  \begin{tabular}{@{}lccccc@{}}
    \toprule
    Method & AUROC$\uparrow$ & $R_{+}\uparrow$ & $R_{-}\uparrow$ & FPR$\downarrow$ &  FAR$_{95}\downarrow$ \\
    \midrule
    Random              & 0.500 & 0.500 & 0.500 & 0.500 & --- \\
    Confidence          & 0.436 & 0.984 & 0.010 & 0.990 &  0.957 \\
    Temporal conf.      & 0.423 & 0.949 & 0.042 & 0.958 &  0.959 \\
    $k$-NN ($k{=}50$)   & 0.617 & 0.661 & 0.495 & 0.505 &  0.874 \\
    \midrule
    Full TAP, $T{=}1$   & 0.679$\pm$0.001 & 0.708$\pm$0.106 & 0.537$\pm$0.123 & 0.463$\pm$0.123 &  0.826$\pm$0.005 \\
    b0+b1, $T{=}1$      & 0.686$\pm$0.001 & \textbf{0.735$\pm$0.083} & 0.527$\pm$0.080 & 0.473$\pm$0.080 &  0.818$\pm$0.009 \\
    Block~1, $T{=}4$    & 0.694$\pm$0.001 & 0.667$\pm$0.108 & 0.608$\pm$0.101 & 0.392$\pm$0.101 &  0.805$\pm$0.019 \\
    \midrule
    Full TAP, $T{=}4$   & 0.692$\pm$0.005 & 0.567$\pm$0.035 & \textbf{0.708$\pm$0.028} & \textbf{0.292$\pm$0.028} &  0.827$\pm$0.017 \\
    \textbf{b0+b1, $T{=}4$} & \textbf{0.704$\pm$0.003} & 0.661$\pm$0.104 & 0.630$\pm$0.089 & 0.370$\pm$0.089 & \textbf{0.759$\pm$0.013} \\
    \bottomrule
  \end{tabular}
\end{table}

The recall columns show that both confidence baselines exceed $0.94$ Error recall only because they flag almost everything, collapsing No-Error recall to $0.010$ and $0.042$. The confidence baselines fall below chance likely because Error frames tend to contain more objects and hence higher maximum scores.

The b0+b1 temporal monitor achieves a higher AUROC, but at a sigmoid threshold of $0.5$ its Error recall is lower than that of the matched frame-only model ($0.661 \pm 0.104$ vs. $0.735 \pm 0.083$). At the same threshold, however, it also reduces the false positive rate from $0.473$ to $0.370$. This indicates that the two models operate at different points on their respective ROC curves rather than one uniformly dominating the other. In deployment, the decision threshold should therefore be selected according to the relative cost of missed errors and false alarms.

For the strongest model (b0+b1), the validation set AUROC is 0.736, a gap of roughly 3.2 points consistent with scene sampling variability between validation and test sets.

\subsection{False Alarm Rate Across Target Recalls}
\label{sec:far_sweep}

Since a deployed monitor would operate at a recall set by the safety requirements, Fig. 2 reports FAR across target recalls. For each seed and target, we select the highest threshold whose empirical TPR meets the target, without interpolation, and average the resulting FAR over seeds.

\begin{figure}[t]
  \centering
  \includegraphics[width=\columnwidth]{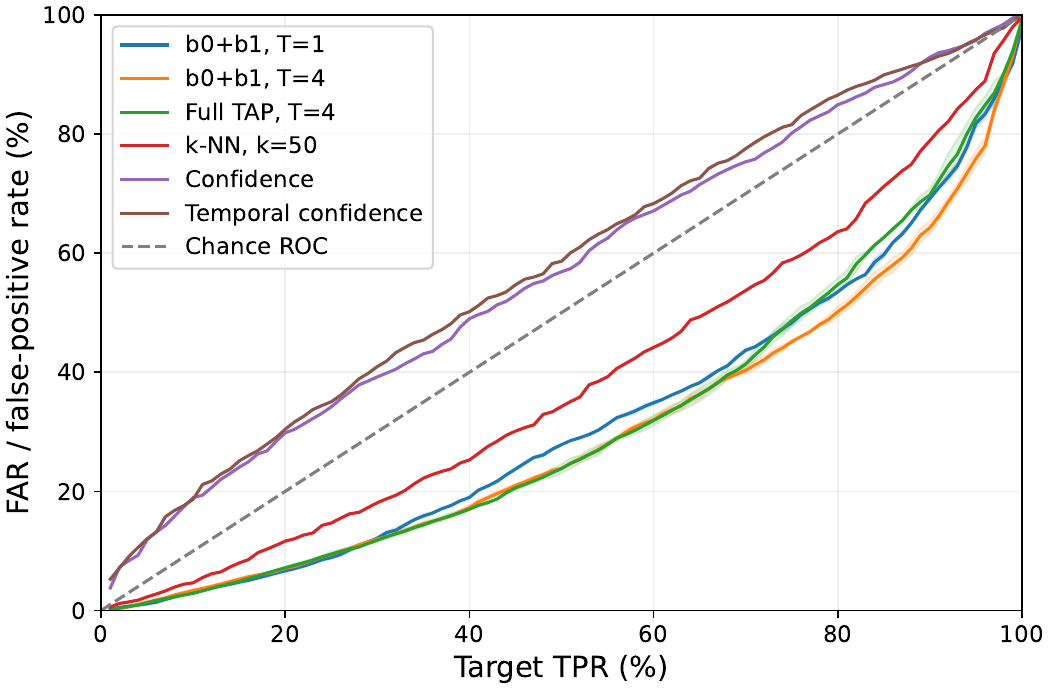}
  \caption{False-alarm rate (FAR) versus target true-positive rate (TPR) on the test set. Curves show empirical ROC operating points evaluated on a one-percentage-point target TPR grid; the nine TPR targets reported in the tables are a subset of these points. Shading indicates one population standard deviation across three seeds for learned monitors. The dashed diagonal represents chance performance.}
  \label{fig:far_curves}
\end{figure}

The b0+b1 temporal monitor attains the lowest FAR of any configuration at every target recall from 70\% to 97\%, requiring a FAR of $0.502$ at 80\% recall against $0.535$ for its matched frame-only counterpart and $0.547$ for the full three-tap fusion. However, at 99\% recall every configuration lies between $0.919$ and $0.946$, where no configuration discriminates well. For the strongest monitor, FAR rises from 0.239 at 50\% recall to 0.642 at 90\% and 0.759 at 95\%.
\subsection{Temporal Window Size Ablation}
\label{sec:window}

The upper block of Table~\ref{tab:temporal} reports performance over three seeds
for $T \in \{1, 2, 4, 6, 8\}$, with all other hyperparameters fixed.
All multi-frame settings outperform the frame-level baseline, with
AUROC forming a shallow plateau between $T{=}2$ and $T{=}8$ spanning
only 0.005. In contrast, FAR@TPR95 decreases from $0.826$ at $T{=}1$ to $0.774$ at $T{=}8$,
indicating that longer temporal context provides a clearer benefit at
high-recall operating points despite the largely unchanged overall
ranking performance.

\begin{table}[t]
  \centering
  \caption{Temporal ablations on the test set (full TAP, $\delta = 3$\,m, 3 seeds). Upper: window $T$ at $h=0$; lower: horizon $h$ at $T=4$.}
  \label{tab:temporal}
  \footnotesize
  \setlength{\tabcolsep}{3pt}
  \begin{tabular}{lcccc}
    \toprule
    Setting  & AUROC$\uparrow$ & $R_{+}\uparrow$ & $R_{-}\uparrow$ & FAR$_{95}\downarrow$ \\
    \midrule
    \multicolumn{5}{l}{\textit{Temporal window $T$ (at $h{=}0$)}} \\
    $T{=}1$ (frame-level)  & 0.679 & 0.708 & 0.537 & 0.826 \\
    $T{=}2$ ($0.5$\,s)     & 0.694 & 0.596 & 0.671 & 0.798 \\
    $T{=}4$ ($1.5$\,s)     & 0.692 & 0.567 & 0.708 & 0.827 \\
    $T{=}6$ ($2.5$\,s)     & 0.697 & 0.608 & 0.672 & 0.781 \\
    $T{=}8$ ($3.5$\,s)     & 0.697 & 0.477 & 0.767 & \textbf{0.774} \\
    \midrule
    \multicolumn{5}{l}{\textit{Prediction horizon $h$ (at $T{=}4$)}} \\
    $h{=}0$ (current)      & 0.692 & 0.567 & 0.708 & 0.827 \\
    $h{=}1$ ($0.5$\,s)     & 0.697 & 0.753 & 0.536 & 0.799 \\
    $h{=}2$ ($1.0$\,s)     & 0.690 & 0.685 & 0.589 & 0.803 \\
    $h{=}3$ ($1.5$\,s)     & 0.687 & 0.717 & 0.576 & 0.822 \\
    $h{=}5$ ($2.5$\,s)     & 0.668 & 0.748 & 0.506 & 0.871 \\
    \bottomrule
  \end{tabular}
\end{table}

\subsection{Prediction Horizon Ablation}
\label{sec:horizon}

The monitor is also evaluated predictively, with the target label shifted forward by $h \in \{0, 1, 2, 3, 5\}$ frames (lower block of Table~\ref{tab:temporal}), while the input remains the $T$ past frames. Performance degrades only modestly as the target moves forward, though not perfectly monotonically: the monitor retains meaningful
discrimination at $h{=}1$ (AUROC $= 0.697$, 0.5\,s ahead at the 2\,Hz annotation rate) and 0.687 at $h{=}3$ (1.5\,s ahead). Each horizon discards scene endpoints with no future target, so the eligible frame count falls from 6{,}019 at $h{=}0$ to 5{,}269 at $h{=}5$ and part of the variation may reflect sample composition rather than task
difficulty; positive prevalence, however, stays stable (15.3--15.8\%). This suggests that the temporal signal has some predictive value, though persistent errors may also contribute.
\subsection{TAP Layer Ablation}
\label{sec:tap_ablation}

Table~\ref{tab:tap} presents AUROC for each of the 7 non-empty
subsets of the TAP layer set $\{$Block~0, Block~1, Neck$\}$.
The strongest configuration is the two-layer combination of Block~0 and
Block~1, indicating that early and mid-level backbone statistics carry
most of the useful signal. The neck alone is the weakest performer by
a clear margin, and every combination including it is worse than the
best combination that excludes it. Fusion is not uniformly beneficial:
Block~1 alone (AUROC $= 0.694$) outperforms both two-layer combinations
with the neck, b0+neck ($0.667$) and b1+neck ($0.685$), and also the
full three-tap fusion ($0.692$). Layer choice therefore matters more than the number of layers; adding the neck increases dimension and cost without benefit.

\begin{table}[t]
  \centering
  \caption{TAP layer ablation on the test set ($T=4$, $h=0$, $\delta = 3$\,m, 3 seeds). Best result in bold.}
  \label{tab:tap}
  \footnotesize
  \setlength{\tabcolsep}{3.5pt}
  \begin{tabular}{lrcccc}
    \toprule
    Taps used & Dim & AUROC$\uparrow$ & $R_{+}\uparrow$ & $R_{-}\uparrow$ & FAR$_{95}\downarrow$ \\
    \midrule
    Block~0            &  640 & 0.661 & 0.597 & 0.607 & 0.838 \\
    Block~1            & 1280 & 0.694 & 0.667 & 0.608 & 0.805 \\
    Neck               & 1280 & 0.637 & 0.561 & 0.648 & 0.898 \\
    \textbf{b0 + b1}   & 1920 & \textbf{0.704} & 0.661 & 0.630 & \textbf{0.759} \\
    b0 + neck          & 1920 & 0.667 & 0.604 & 0.615 & 0.849 \\
    b1 + neck          & 2560 & 0.685 & 0.804 & 0.438 & 0.831 \\
    b0 + b1 + neck     & 3200 & 0.692 & 0.567 & 0.708 & 0.827 \\
    \bottomrule
  \end{tabular}
\end{table}

\subsection{Computational Cost}
\label{sec:compute}

At inference, \method{} adds three components to the base CenterPoint
detector: activation extraction through forward hooks, TAP-Stats
encoding, and temporal inference. Table~\ref{tab:compute} reports the
standalone complexity of the temporal head for the main configurations.
The three-tap configuration ($d{=}3200$, $T{=}4$) requires 2.31\,MFLOPs
and 582{,}657 parameters at a median CPU latency of 0.174\,ms; the
lighter b0+b1 configuration, the strongest model, reduces this
to 1.65\,MFLOPs, 418{,}817 parameters, and 0.159\,ms. Moving from
$T{=}1$ to $T{=}4$ costs only 0.052\,ms at the same parameter count, so
the temporal gain in Section~\ref{sec:main_results} is obtained at
negligible additional cost. These figures are a standalone profile of
the monitor head alone, excluding the detector, forward hooks,
TAP-Stats encoding, and host-device transfers.

\begin{table}[t]
  \centering
  \caption{Standalone complexity of the temporal monitor head. Latency
is median CPU inference time, excluding activation extraction and
TAP-Stats encoding.}
  \label{tab:compute}
  \footnotesize
  \setlength{\tabcolsep}{4pt}
  \begin{tabular}{lcccc}
    \toprule
    Variant & Dim & Params & FLOPs & Latency (ms) \\
    \midrule
    b0+b1, $T{=}1$       & 1920 & 418.8K & 0.42M & 0.107 \\
    b0+b1, $T{=}4$       & 1920 & 418.8K & 1.65M & 0.159 \\
    3-block, $T{=}4$     & 3200 & 582.7K & 2.31M & 0.174 \\
    3-block, $T{=}8$     & 3200 & 582.7K & 4.61M & 0.188 \\
    \bottomrule
  \end{tabular}
\end{table}

\section{Discussion}

The results show that temporal monitoring improves failure discrimination over the matched frame-level monitor. Most of the benefit comes from short-term context rather than long-range dependencies (a 0.005 AUROC plateau from $T{=}2$ to $T{=}8$), and activation trajectories carry information about near-future failures, with discrimination at $h{=}1$ matching $h{=}0$, and degrading modestly up to 1.5 s.

Representation choice also matters as the strongest configuration uses Block~0 and Block~1 rather than the full three-tap fusion. Block~1 alone outperforms both neck-containing pairs and the full representation, since the neck adds dimension and cost without improving AUROC or high-recall behaviour.

These findings position \method{} as a low-cost supervisory signal rather than a standalone safety mechanism: the false-alarm rate rises steeply above 90\% recall and is close to indiscriminate by 99\%, so its output is better suited to complementing other evidence within a broader diagnostic or fallback architecture than to standalone alarming.

\textbf{Limitations:} evaluation is restricted to NuScenes and a single detector; the frame-level label does not identify \textit{which} object was missed, so the horizon results predict future labels rather than an event-onset warning time; FAR@TPR95 remains substantial (0.759 for the strongest model); reported standard deviations reflect seed rather than scene-sampling variability; and an auxiliary sweep of the missed-object matching threshold (1--4\,m, not reported in full here) shows AUROC rising with a stricter threshold while FAR@TPR95 stays roughly flat, indicating sensitivity to task definition.

\section{Conclusion}
\label{sec:conclusion}

This paper presented \method{}, a temporal self-assessment mechanism for
run-time integrity monitoring of LiDAR-based 3D object detection in
automated driving systems. Results show that incorporating
temporal context into activation pattern monitoring yields measurable
and consistent improvements in false-negative detection. By extracting
five-statistic summaries of intermediate CenterPoint activations and
processing them with a TCN, \method{} outperforms confidence-based and frame-level baselines on NuScenes: on
the default 3\,m task, the strongest configuration (b0+b1 taps)
achieves AUROC~$= 0.704 \pm 0.003$ and
FAR@TPR95~$= 0.759 \pm 0.013$, leading every evaluated configuration on
both measures.

Ablations on the test set show that all multi-frame windows outperform frame-only inputs. Early and mid-level backbone taps are most informative, with the best results from combining Block 0 and 1, while a lightweight single-block TCN suffices for temporal aggregation. The resulting monitor head is computationally cheap, but the false-alarm rates at high recall indicate that practical deployment remains an open challenge. Hence, future work will
investigate more expressive temporal architectures, extend the horizon analysis from frame labels to event onsets, evaluate generalisation across additional datasets and detectors, and explore fusing camera- and LiDAR-derived monitoring signals.


\bibliographystyle{IEEEtran}
\bibliography{references}


\end{document}